\documentclass[Afour,sageh,times]{sagej}

\usepackage{moreverb,url}

\usepackage{graphicx}
\usepackage{amsmath}
\usepackage{amssymb}
\usepackage{color}
\usepackage{multirow}
\usepackage[dvipsnames]{xcolor}

\newcommand{\ie}{\textit{i}.\textit{e}. }
\newcommand{\eg}{\textit{e}.\textit{g}. }
\newcommand{\mypara}{\par\vspace*{0.8mm}\noindent\textbf}
\newcommand{\myedit}{\textcolor{black}}
\newcommand{\mysecondedit}{\textcolor{black}}

\usepackage[colorlinks,bookmarksopen,bookmarksnumbered,citecolor=red,urlcolor=red]{hyperref}

\newcommand\BibTeX{{\rmfamily B\kern-.05em \textsc{i\kern-.025em b}\kern-.08em
T\kern-.1667em\lower.7ex\hbox{E}\kern-.125emX}}

\begin{document}

\runninghead{Zeng et al.}

\title{\hspace{-7.5mm}\textrm{Robotic Pick-and-Place of Novel Objects\\\hspace{-7.5mm}in Clutter with Multi-Affordance Grasping\\\hspace{-7.5mm}and Cross-Domain Image Matching}}

\author{\textrm{Andy Zeng\affilnum{1}, 
Shuran Song\affilnum{1}, 
Kuan-Ting Yu\affilnum{2}, 
Elliott Donlon\affilnum{2}, 
\\Francois R. Hogan\affilnum{2}, 
Maria Bauza\affilnum{2}, 
Daolin Ma\affilnum{2}, 
Orion Taylor\affilnum{2}, 
\\Melody Liu\affilnum{2}, 
Eudald Romo\affilnum{2}, 
Nima Fazeli\affilnum{2}, 
Ferran Alet\affilnum{2}, 
\\Nikhil Chavan Dafle\affilnum{2}, 
Rachel Holladay\affilnum{2}, 
Isabella Morona\affilnum{2}, 
\\Prem Qu Nair\affilnum{1}, 
Druck Green\affilnum{2}, 
Ian Taylor\affilnum{2}, 
Weber Liu\affilnum{1}, 
\\Thomas Funkhouser\affilnum{1}, 
Alberto Rodriguez\affilnum{2} 
}
}

\affiliation{\affilnum{1}Princeton University, NJ, USA\\
\affilnum{2}Massachusetts Institute of Technology, Cambridge, MA, USA}

\corrauth{Andy Zeng, Princeton University, NJ, USA}
\email{andyz@cs.princeton.edu}

\begin{abstract}
This paper presents a robotic pick-and-place system that is capable of grasping and recognizing both known and novel objects in cluttered environments. The key new feature of the system is that it handles a wide range of object categories without needing any task-specific training data for novel objects. To achieve this, it first uses an object-agnostic grasping framework to map from visual observations to actions: inferring dense pixel-wise probability maps of the affordances for four different grasping primitive actions. It then executes the action with the highest affordance and recognizes picked objects with a cross-domain image classification framework that matches observed images to product images. Since product images are readily available for a wide range of objects (e.g., from the web), the system works out-of-the-box for novel objects without requiring any additional data collection or re-training. 
Exhaustive experimental results demonstrate that our multi-affordance grasping achieves high success rates for a wide variety of objects in clutter, and our recognition algorithm achieves high accuracy for both known and novel grasped objects. The approach was part of the MIT-Princeton Team system that took 1st place in the stowing task at the 2017 Amazon Robotics Challenge. All code, datasets, and pre-trained models are available online at \href{http://arc.cs.princeton.edu/}{http://arc.cs.princeton.edu}
\end{abstract}


\maketitle
{\let\thefootnote\relax\footnote{{This paper is a revision of a paper appearing in the proceedings of the 2018 International Conference on Robotics and Automation~\cite{zeng2018robotic}.}}}
\vspace{-0.5in}

\section{Introduction}

A human's remarkable ability to grasp and recognize unfamiliar objects with little prior knowledge of them is a constant inspiration for robotics research.
This ability to grasp the unknown is central to many applications: from picking packages in a logistic center to bin-picking in a manufacturing plant; from unloading groceries at home to clearing debris after a disaster.
The main goal of this work is to demonstrate that it is possible -- and practical -- for a robotic system to pick and recognize novel objects with very limited prior information about them (e.g. with only a few representative images scraped from the web).

Despite the interest of the research community, and despite its practical value, robust manipulation and recognition of novel objects in cluttered environments still remains a largely unsolved problem.
Classical solutions for robotic picking require recognition and pose estimation prior to model-based grasp planning, or require object segmentation to associate grasp detections with object identities. These solutions tend to fall short when dealing with novel objects in cluttered environments, since they rely on 3D object models that are not available and/or large amounts of training data to achieve robust performance. Although there has been inspiring recent work on detecting grasps directly from RGB-D pointclouds as well as learning-based recognition systems to handle the constraints of novel objects and limited data, these methods have yet to be proven in the constraints and accuracy required by a real task with heavy clutter, severe occlusions, and object variability. 

\begin{figure}[t]
\centering
  \vspace{2mm}
  \includegraphics[width=\linewidth]{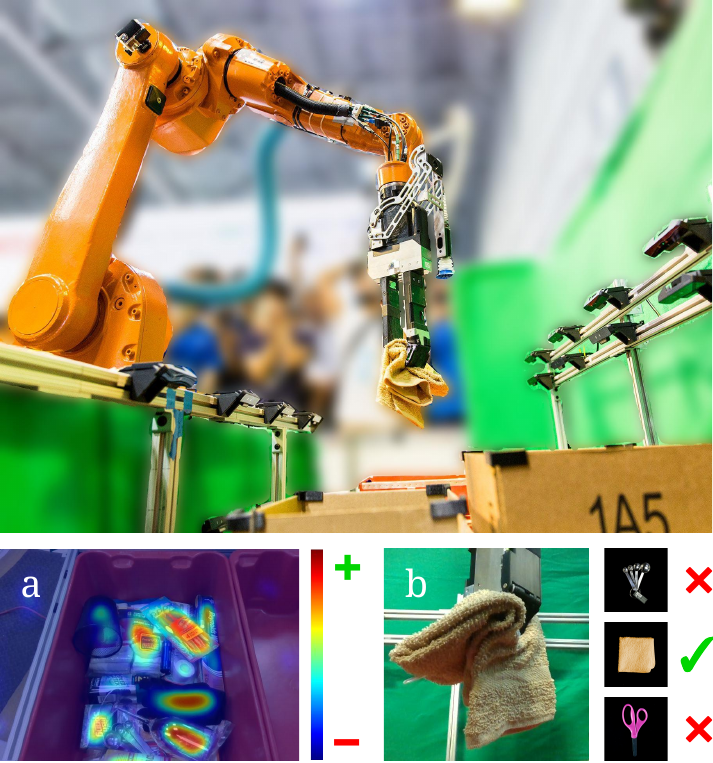}
  \caption{{\bf Our picking system} computing pixel-wise affordances for grasping over visual observations of bins full of objects, (a) grasping a towel and holding it up away from clutter, and recognizing it by matching observed images of the towel (b) to an available representative product image. The key contribution is that the entire system works out-of-the-box for novel objects (unseen in training) without the need for any additional data collection or re-training.
  }
  \vspace{-2mm}
  \label{fig:teaser}
\end{figure}

In this paper, we propose a system that picks and recognizes objects in cluttered environments. We have designed the system specifically to handle a wide range of objects novel to the system without gathering any task-specific training data from them.  

To make this possible, our system consists of two components. The first is a multi-affordance grasping framework which uses fully convolutional networks (FCNs) to take in visual observations of the scene and output dense predictions (arranged with the same size and resolution as the input data) measuring the affordance (or probability of picking success) for four different grasping primitive actions over a pixel-wise sampling of end-effector orientations and locations. The primitive action with the highest inferred affordance value determines the picking action executed by the robot. This picking framework operates without a priori object segmentation and classification and hence is agnostic to object identity. 

The second component of the system is a cross-domain image matching framework for recognizing grasped objects by matching them to product images using a two-stream convolutional network (ConvNet) architecture. This framework adapts to novel objects without additional re-training.
Both components work hand-in-hand to achieve robust picking performance of novel objects in heavy clutter.

We provide exhaustive experiments, ablation, and comparison to evaluate both components. We demonstrate that our affordance-based algorithm for grasp planning achieves high success rates for a wide variety of objects in clutter, and the recognition algorithm achieves high accuracy for known and novel grasped objects.
These algorithms were developed as part of the MIT-Princeton Team system that took 1st place in the stowing task of the Amazon Robotics Challenge (ARC), being the only system to have successfully stowed all known and novel objects from an unstructured tote into a storage system within the allotted time frame. Figure~\ref{fig:teaser} shows our robot in action during the competition.

In summary, our main contributions are:
\begin{itemize}
    \item{An affordance-based object-agnostic perception framework to plan grasps using four primitive grasping actions for fast and robust picking. This utilizes fully convolutional networks for inferring dense pixel-wise affordances of each primitive (Section \ref{sec:manipulation}).} 
    \item{A perception framework for recognizing both known and novel objects using only product images without extra data collection or re-training. This utilizes a two stream convolutional network to match images or picked objects to product images (Section \ref{sec:recognition}).}
    \item{A system combining these two frameworks for picking novel objects in heavy clutter.} 
\end{itemize}
All code, datasets, and pre-trained models are available online at \href{http://arc.cs.princeton.edu/}{http://arc.cs.princeton.edu}. We also provide a video summarizing our approach at \href{https://youtu.be/6fG7zwGfIkI}{https://youtu.be/6fG7zwGfIkI}.

\section{Related Work}

In this section, we review works related to robotic picking systems. Works specific to grasping (Section \ref{sec:manipulation}) and recognition (Section \ref{sec:recognition}) are in their respective sections.

\subsection{Recognition followed by Model-based Grasping}

A large number of autonomous pick-and-place solutions follow a standard two-step approach: object recognition and pose estimation followed by model-based grasp planning. For example, \cite{jonschkowski2016probabilistic} designed object segmentation methods over handcrafted image features to compute suction proposals for picking objects with a vacuum. 

More recent data-driven approaches ( \cite{hernandez2016team,zeng2016multi,schwarz2017nimbro,wong2017segicp}) use ConvNets to provide bounding box proposals or segmentations, followed by geometric registration to estimate object poses, which ultimately guide handcrafted picking heuristics (\cite{Bicchi2000RoboticContact,Miller2003}). \cite{nieuwenhuisen2013mobile} improve many aspects of this pipeline by leveraging robot mobility, while \cite{liu2012fast} adds a pose correction stage when the object is in the gripper. These works typically require 3D models of the objects during test time, and/or training data with the physical objects themselves. This is practical for tightly constrained pick-and-place scenarios, but is not easily scalable to applications that consistently encounter novel objects, for which only limited data (i.e. product images from the web) is available.

\subsection{Recognition in parallel with Object-Agnostic Grasping}

It is also possible to exploit local features of objects without object identity to efficiently detect grasps (\cite{morales2004, lenz2015rss,redmon2015, ten2015using, pinto2016, pinto2017, mahler2017, gualtieri2017,levine2016learning}). Since these methods are agnostic to object identity, they better adapt to novel objects and experience higher picking success rates in part by eliminating error propagation from a prior recognition step. \cite{matsumotoend} apply this idea in a full picking system by using a ConvNet to compute grasp proposals, while in parallel inferring semantic segmentations for a fixed set of known objects. 
Although these pick-and-place systems use object-agnostic grasping methods, they still require some form of in-place object recognition in order to associate grasp proposals with object identities, which is particularly challenging when dealing with novel objects in clutter.







\begin{figure}[t]
\centering
  \includegraphics[width=\linewidth]{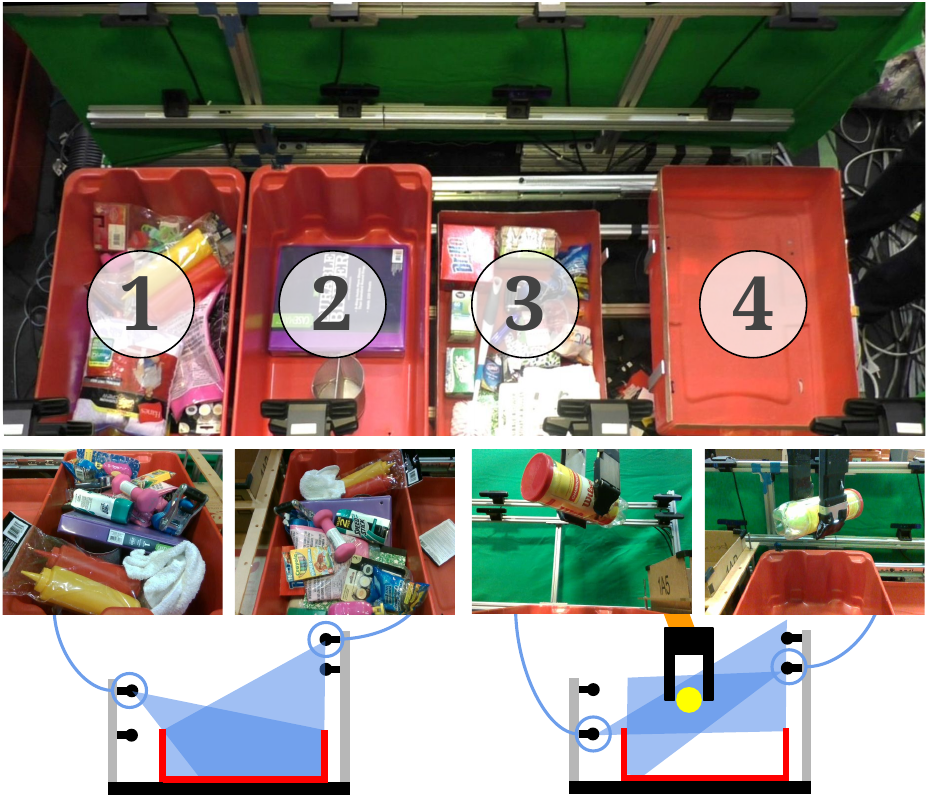}
  \caption{{\bf The bin and camera setup.} Our system consists of 4 units (top), where each unit has a bin with 4 stationary cameras: two overlooking the bin (bottom-left) are used for inferring grasp affordances while the other two (bottom-right) are used for recognizing grasped objects.}
  \label{fig:setup-wide}
  \vspace{-3mm}
\end{figure}

\subsection{Active Perception}

The act of exploiting control strategies for acquiring data to improve perception (\cite{bajcsy1992active,chen2011active}) can facilitate the recognition of novel objects in clutter.
For example, \cite{jiang2016novel} describe a robotic system that actively rearranges objects in the scene (by pushing) in order to improve recognition accuracy. Other works \cite{wu2015active,jayaraman2016look} explore next-best-view based approaches to improve recognition, segmentation and pose estimation results. 
Inspired by these works, our system uses a form of active perception by using a grasp-first-then-recognize paradigm where we leverage object-agnostic grasping to isolate each object from clutter in order to significantly improve recognition accuracy for novel objects. 



\section{System Overview}

We present a robotic pick-and-place system that grasps and recognizes both known and novel objects in cluttered environments. 
We will refer by ``known'' objects to those that are provided to the system at training time, both as physical objects and as representative product images (images of objects available on the web); while ``novel'' objects are provided only at test time in the form of representative product images. 

The pick-and-place task presents us with two main perception challenges: 1) find accessible grasps of objects in clutter; and 2) match the identity of grasped objects to product images. Our approach and contributions to these two challenges are described in detail in Section~\ref{sec:manipulation} and Section~\ref{sec:recognition} respectively.
For context, in this section we briefly describe the system that will use those two capabilities.


\mypara{Overall approach.} The system follows a \textit{grasp-first-then-recognize} work-flow. For each pick-and-place operation, it first uses FCNs to infer the pixel-wise affordances of four different grasping primitive actions: from suction to parallel-jaw grasps (Section \ref{sec:manipulation}). It then selects the grasping primitive action with the highest affordance, picks up one object, isolates it from the clutter, holds it up in front of cameras, recognizes its category, and places it in the appropriate bin. Although the object recognition algorithm is trained only on known objects, it is able to recognize novel objects through a learned cross-domain image matching embedding between observed images of held objects and product images (Section \ref{sec:recognition}). 

\mypara{Advantages.} This system design has several advantages.
First, the affordance-based grasping algorithm is model-free and agnostic to object identities and generalizes to novel objects without re-training. 
Second, the category recognition algorithm works without task-specific data collection or re-training for novel objects, which makes it scalable for applications in warehouse automation and service robots where the range of observed object categories is large and dynamic.   
Third, our grasping framework supports multiple grasping modes with a multi-functional gripper and thus handles a wide variety of objects.
Finally, the entire processing pipeline requires only a few forward passes through deep networks and thus executes quickly \myedit{(run-times reported in Table \ref{table:speed})}.
\begin{figure}[t]
\centering
  \includegraphics[width=\linewidth]{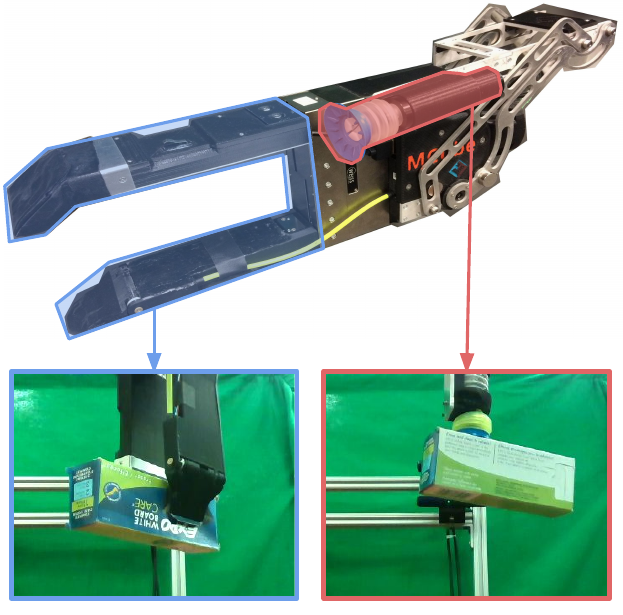}
  \caption{{\bf Multi-functional gripper} with a retractable mechanism that enables quick and automatic switching between suction (pink) and grasping (blue).}
  \label{fig:gripper}
  \vspace{-3mm}
\end{figure}

\begin{figure*}[t]
\centering
  \includegraphics[width=\linewidth]{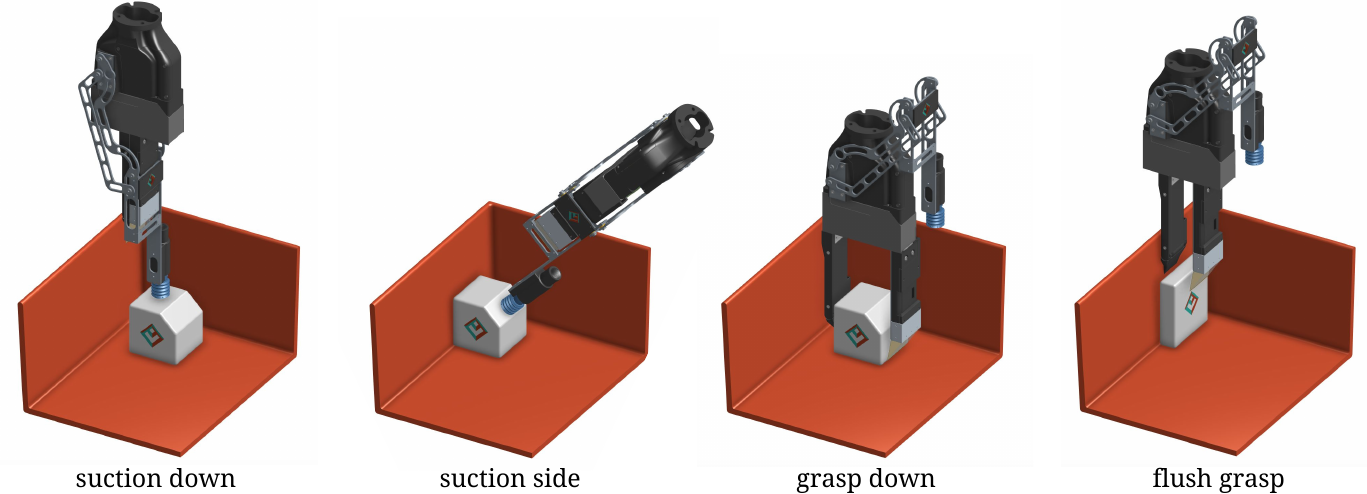}
  \caption{{\bf Multiple motion primitives} for suction and grasping to ensure successful picking for a wide variety of objects in any orientation.}
  \label{fig:primitives}
  \vspace{-3mm}
\end{figure*}

\mypara{System setup.} Our system features a 6DOF ABB IRB 1600id robot arm next to four picking work-cells.  The robot arm's end-effector is a multi-functional gripper with two fingers for parallel-jaw grasps and a retractable suction cup (Fig.~\ref{fig:gripper}).  This gripper was designed to function in cluttered environments: finger and suction cup length are specifically chosen such that the bulk of the gripper body does not need to enter the cluttered space.

Each work-cell has a storage bin and four statically-mounted RealSense SR300 RGB-D cameras (Fig.\ref{fig:setup-wide}): two cameras overlooking the storage bins are used to infer grasp affordances, while the other two pointing upwards towards the robot gripper are used to recognize objects in the gripper. \myedit{For the two cameras used to infer grasp affordances, we find that placing them at opposite viewpoints of the storage bins provides good visual coverage of the objects in the bin. Adding a third camera did not significantly improve visual coverage. For the other two cameras used for object recognition, having them at opposite viewpoints enables us to immediately reconstruct a near-complete 3D point cloud of the object while it is being held in the gripper. These 3D point clouds are useful for planning object placements in the storage system.}

Although our experiments were performed with this setup, the system was designed to be flexible for picking and placing between any number of reachable work-cells and camera locations.  Furthermore, all manipulation and recognition algorithms in this paper were designed to be easily adapted to other system setups.

\section{Challenge I: Planning Grasps with Multi-Affordance Grasping}
\label{sec:manipulation}
The goal of the first step in our system is to robustly grasp objects from a cluttered scene without relying on their object identities or poses. To this end, we define a set of four grasping primitive actions that are complementary to each other in terms of utility across different object types and scenarios -- empirically broadening the variety of objects and orientations that can be picked with at least one primitive. Given RGB-D images of the cluttered scene at test time, we infer the dense pixel-wise affordances for all four primitives. A task planner then selects and executes the primitive with the highest affordance. 

\subsection{Grasping Primitives}

We define four grasping primitives to achieve robust picking for typical household objects. Figure~\ref{fig:primitives} shows example motions for each primitive. Each of them is implemented as a set of guarded moves with collision avoidance using force sensors below the work-cells. They also have quick success or failure feedback mechanisms using either flow sensing for suction or force sensing for grasping. Robot arm motion planning is automatically executed within each primitive with stable inverse kinematic-based controllers~\cite{diankov_thesis}. These primitives are as follows:

\mypara{Suction down} grasps objects with a vacuum gripper vertically. This primitive is particularly robust for objects with large and flat suctionable surfaces (e.g. boxes, books, wrapped objects), and performs well in heavy clutter.

\mypara{Suction side} \myedit{grasps objects from the side by approaching with a vacuum gripper tilted at a fixed angle.} This primitive is robust to thin and flat objects resting against walls, which may not have suctionable surfaces from the top.

\mypara{Grasp down} grasps objects vertically using the two-finger parallel-jaw gripper. This primitive is complementary to the suction primitives in that it is able to pick up objects with smaller, irregular surfaces (e.g. small tools, deformable objects), or made of semi-porous materials that prevent a good suction seal (e.g. cloth).

\mypara{Flush grasp} retrieves unsuctionable objects that are flushed against a wall. The primitive is similar to grasp down, but with the additional behavior of using a flexible spatula to slide one finger in between the target object and the wall.


\begin{figure*}[t]
\centering
  \includegraphics[width=\linewidth]{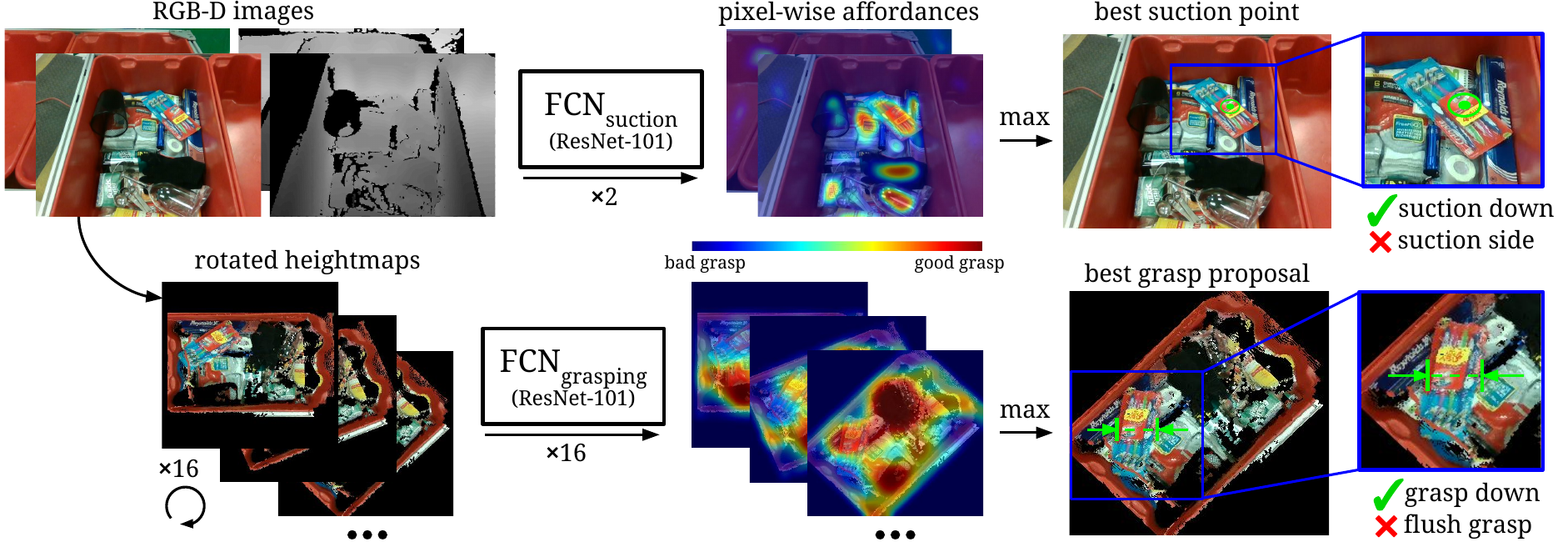}
  \caption{{\bf Learning pixel-wise affordances for suction and grasping.} Given multi-view RGB-D images, we infer pixel-wise suction affordances for each image with an FCN (top row). The inferred affordance value at each pixel describes the utility of suction at that pixel's projected 3D location. We aggregate the inferred affordances onto a 3D point cloud, where each point corresponds to a suction proposal (down or side based on surface normals). In parallel, we merge RGB-D images into an orthographic RGB-D heightmap of the scene, rotate it by 16 different angles, and feed them each through another FCN (bottom row) to estimate the pixel-wise affordances of horizontal grasps for each heightmap. This effectively produces affordance maps for 16 different top-down grasping angles, from which we generate grasp down and flush grasp proposals. The suction or grasp proposal with the highest affordance value is executed.}
  \label{fig:passive-network}
  \vspace{-3mm}
\end{figure*}

\subsection{Learning Affordances with Fully Convolutional Networks}

Given the set of pre-defined grasping primitives and RGB-D images of the scene, we train FCNs (\cite{long_shelhamer_fcn}) to infer the affordances for each primitive across a dense pixel-wise sampling of end-effector orientations and locations (\textit{i.e.} each pixel correlates to a different position on which to execute the primitive). 
%
Our approach relies on the assumption that graspable regions can be deduced from local geometry and visual appearance. This is inspired by recent data-driven methods for grasp planning \cite{morales2004, saxena2008robotic, lenz2015rss,redmon2015, pinto2016, pinto2017, mahler2017, gualtieri2017,levine2016learning}, which do not rely on object identities or state estimation. 


\mypara{Inferring Suction Affordances.} We define suction points as 3D positions where the vacuum gripper's suction cup should come in contact with the object's surface in order to successfully grasp it. Good suction points should be located on suctionable (\eg nonporous) surfaces, and nearby the target object's center of mass to avoid an unstable suction seal (e.g. particularly for heavy objects). Each suction proposal is defined as a suction point, its local surface normal (computed from the projected 3D point cloud), and its affordance value. Each pixel of an RGB-D image (with a valid depth value) maps surjectively to a suction point.

We train a fully convolutional residual network (ResNet-101 \cite{he2016deep}), that takes a $640\times480$ RGB-D image as input, and outputs a densely labeled pixel-wise map (with the same image size and resolution as the input) of affordance values between 0 and 1. Values closer to one imply a more preferable suction location. Visualizations of these densely labeled affordance maps are shown as heat maps in the first row of Fig.~\ref{fig:passive-network}. Our network architecture is multi-modal, where the color data (RGB) is fed into one ResNet-101 tower, and 3-channel depth (DDD, cloned across channels, normalized by subtracting mean and dividing by standard deviation) is fed into another ResNet-101 tower. \myedit{The depth is cloned across channels so that we can use the ResNet weights pre-trained on 3-channel (RGB) color images from ImageNet \cite{deng2009imagenet} to process depth information.} Features from the ends of both towers are concatenated across channels, followed by 3 additional spatial convolution layers to merge the features; then spatially bilinearly upsampled and softmaxed to output a binary probability map representing the inferred affordances.

Our FCN is trained over a manually annotated dataset of RGB-D images of cluttered scenes with diverse objects, where pixels are densely labeled either positive, negative, or neither. Pixel regions labeled as neither are trained with 0 loss backpropagation. We train our FCNs by stochastic gradient descent with momentum, using fixed learning rates of $10^{-3}$ and momentum of 0.99. Our models are trained in Torch/Lua with an NVIDIA Titan X on an Intel Core i7-3770K clocked at 3.5 GHz. Training takes about 10 hours.

During testing, we feed each captured RGB-D image through our trained network to generate dense suction affordances for each view of the scene. As a post-processing step, we use calibrated camera intrinsics and poses to project the RGB-D data and aggregate the affordances onto a combined 3D point cloud. We then compute surface normals for each 3D point (using a local region around it), which are used to classify which suction primitive (down or side) to use for the point.

To handle objects that lack depth information, e.g., finely meshed objects or transparent objects, we use a simple hole filling algorithm \cite{NYUdataset} on the depth images, and project inferred affordance values onto the hallucinated depth. We filter out suction points from the background by performing background subtraction \cite{zeng2016multi} between the captured RGB-D image of the scene with objects and an RGB-D image of the scene without objects (captured automatically before any objects are placed into the picking work-cells).

\mypara{Inferring Grasp Affordances.} Grasp proposals are represented by 1) a 3D position which defines the middle point between the two fingers during top-down parallel-jaw grasping, 2) an angle which defines the orientation of the gripper around the vertical axis along the direction of gravity, 3) the width between the gripper fingers during the grasp, and 4) its affordance value. 

Two RGB-D views of the scene are aggregated into a registered 3D point cloud, which is then orthographically back-projected upwards in the gravity direction to obtain a ``heightmap" image representation of the scene with both color (RGB) and height-from-bottom (D) channels. Each pixel of the heightmap represents a $2x2$mm vertical column of 3D space in the scene. Each pixel also correlates bijectively to a grasp proposal whose 3D position is naturally computed from the spatial 2D position of the pixel relative to the heightmap image and the height value at that pixel. The gripper orientation of the grasp proposal is always kept horizontal with respect to the frame of the heightmap. 

Analogous to our deep network inferring suction affordances, we feed this RGB-D heightmap as input to a fully convolutional ResNet-101 \cite{he2016deep}, which densely infers affordance values (between 0 and 1) for each pixel -- thereby for all top-down parallel-jaw grasping primitives executed with a horizontally orientated gripper across all 3D locations in heightmap of the scene sampled at pixel resolution. Visualizations of these densely labeled affordance maps are shown as heat maps in the second row of Fig.~\ref{fig:passive-network}. By rotating the heightmap of the scene with $n$ different angles prior to feeding as input to the FCN, we can account for $n$ different gripper orientations around the vertical axis. For our system $n=16$; hence we compute affordances for all top-down parallel-jaw grasping primitives with $16$ forward passes of our FCN to generate $16$ output affordance maps. 

We train our FCN over a manually annotated dataset of RGB-D heightmaps, where each positive and negative grasp label is represented by a pixel on the heightmap as well as an angle indicating the preferred gripper orientation. We trained this FCN with the same optimization parameters as that of the FCN used for inferring suction affordances.

During post-processing, the width between the gripper fingers for each grasp proposal is determined by using the local geometry of the 3D point cloud. We also use the location of each proposal relative to the bin to classify which grasping primitive (down or flush) should be used: flush grasp is executed for pixels located near the sides of the bins; grasp down is executed for all other pixels. To handle objects without depth, we triangulate no-depth regions in the heightmap using both RGB-D camera views of the scene, and fill in these regions with synthetic height values of 3cm prior to feeding into the FCN. We filter out inferred grasp proposals in the background by using background subtraction with the RGB-D heightmap of an empty work-cell.






\subsection{Other Architectures for Parallel-Jaw Grasping}

A significant challenge during the development of our system was designing a deep network architecture for inferring dense affordances for parallel-jaw grasping that 1) supports various gripper orientations and 2) could converge during training with less than 2000 manually labeled images. It took several iterations of network architecture designs before discovering the one that worked (described above). Here, we briefly review some deprecated architectures and their primary drawbacks:

\textbf{Parallel trunks and branches ($n$ copies).} This design consists of $n$ separate FCNs, each responsible for inferring the output affordances for one of $n$ grasping angles. Each FCN shares the same architecture: a multi-modal trunk (with color (RGB) and depth (DDD) data fed into two ResNet-101 towers pre-trained on ImageNet, where features at the ends of both towers are concatenated across channels), followed by 3 additional spatial convolution layers to merge the features; then spatially bilinearly upsampled and softmaxed to output an affordance map. This design is similar to our final network design, but with two key differences: 1) there are multiple FCNs, one for each grasping angle, and 2) the input data is not rotated prior to feeding as input to the FCNs. This design is sample inefficient, since each network during training is optimized to learn a different set of visual features to support a specific grasping angle, thus requiring a substantial amount of training samples with that specific grasping angle to converge. Our small manually annotated dataset is characterized by an unequal distribution of training samples across different grasping angles, some of which have as little as less than 100 training samples. Hence, only a few of the FCNs (for grasping angles of which have more than 1,000 training samples) are able to converge during training. Furthermore, attaining the capacity to pre-load all $n$ FCNs into GPU memory for test time requires multiple GPUs.

\textbf{One trunk, split to $n$ parallel branches.} This design consists of a single FCN architecture, which contains a multi-modal ResNet-101 trunk followed by a split into $n$ parallel, individual branches, one for each grasping angle. Each branch contains 3 spatial convolution layers followed by spatial bilinearly upsampling and softmax to output affordance maps. While more lightweight in terms of GPU memory consumption (\ie the trunk is shared and only the 3-layer branches have multiple copies), this FCN still runs into similar training convergence issues as the previous architecture, where each branch during training is optimized to learn a different set of visual features to support a specific grasping angle. The uneven distribution of limited training samples in our dataset made it so that only a few branches are able to converge during training.

\textbf{One trunk, rotate, one branch.} This design consists of a single FCN architecture, which contains a multi-modal ResNet-101 trunk, followed by a spatial transform layer \cite{jaderberg2015spatial} to rotate the intermediate feature map from the trunk with respect to an input grasp angle (such that the gripper orientation is aligned horizontally to the feature map), followed by a branch with 3 spatial convolution layers, spatially bilinearly upsampled, and softmaxed to output a single affordance map for the input grasp angle. This design is even more lightweight than the previous architecture in terms of GPU memory consumption, performs well with grasping angles for which there is a sufficient amount of training samples, but continues to performs poorly for grasping angles with very few training samples (less than 100).

\textbf{One trunk and branch (rotate $n$ times).} This is the final network architecture design as proposed above, which differs from the previous design in that the rotation occurs directly on the input image representation prior to feeding through the FCN (rather than in the middle of the architecture). This enables the entire network to share visual features across different grasping orientations, enabling it to generalize for grasping angles of which there are very few training samples.

\begin{figure*}
\centering
\includegraphics[width=\linewidth]{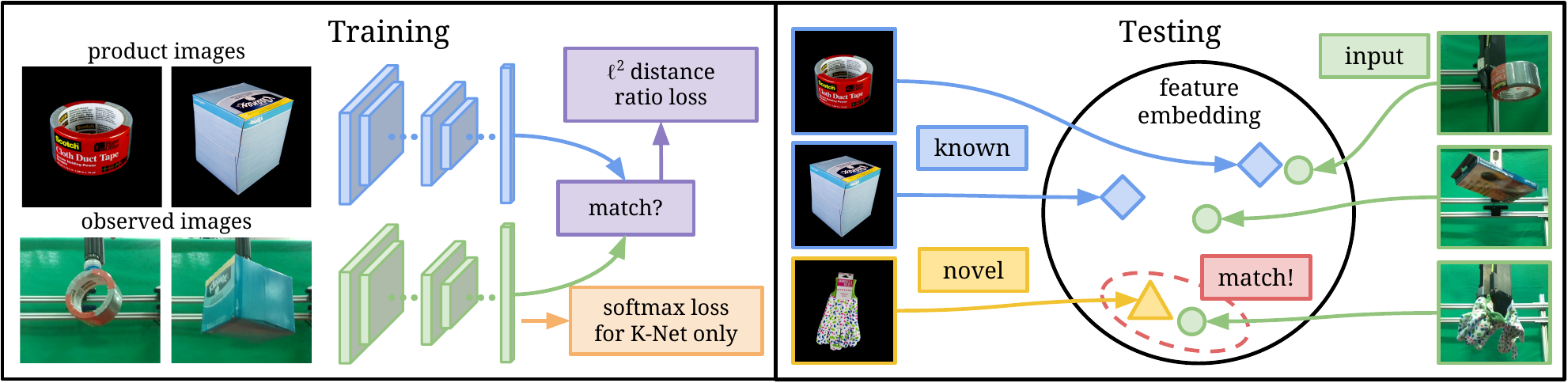}
\caption{{\bf Recognition framework for novel objects.} We train a two-stream convolutional neural network where one stream computes 2048-dimensional feature vectors for product images while the other stream computes 2048-dimensional feature vectors for observed images, and optimize both streams so that features are more similar for images of the same object and dissimilar otherwise. 
During testing, product images of both known and novel objects are mapped onto a common feature space. We recognize observed images by mapping them to the same feature space and finding the nearest neighbor match.}\label{fig:active-vision}
\vspace{-4mm}
\end{figure*}

\subsection{Task Planner}

Our task planner selects and executes the suction or grasp proposal with the highest affordance value. Prior to this, affordance values are scaled by a factor $\gamma_\psi$ that is specific to the proposals' primitive action types $\psi\in\{\mathrm{sd,ss,gd,fg}\}$: suction down (sd), suction side (ss), grasp down (gd), or flush grasp (fg). The value of $\gamma_\psi$ is determined by several task-specific heuristics that induce more efficient picking under competition settings at the ARC. Here we briefly describe these heuristics:

\mypara{Suction first, grasp later.} We empirically find suction to be more reliable than parallel-jaw grasping when picking in scenarios with heavy clutter (10+ objects). Among several factors, the key reason is that suction is significantly less intrusive than grasping. Hence, to reflect a greedy picking strategy that initially favors suction over grasping, $\gamma_\mathrm{gd} = 0.5$ and $\gamma_\mathrm{fg} = 0.5$ for the first 3 minutes of either ARC task (stowing or picking).

\mypara{Avoid repeating unsuccessful attempts.} It is possible for the system to get stuck repeatedly executing the same (or similar) suction or grasp proposal as no change is made to the scene (and hence affordance estimates remain the same). Therefore, after each unsuccessful suction or parallel-jaw grasping attempt, the affordances of the proposals (for the same primitive action) nearby within radius 2cm of the unsuccessful attempt are set to 0.

\mypara{Encouraging exploration upon repeat failures.} The planner re-weights grasping primitive actions $\gamma_\psi$ depending on how often they fail. For primitives that have been unsuccessful for two times in the last 3 minutes, $\gamma_\psi = 0.5$; if unsuccessful for more than three times, $\gamma_\psi = 0.25$. This not only helps the system avoid repeating unsuccessful actions, but also prevents it from excessively relying on any one primitive that doesn't work as expected (\textit{e.g.} in the case of an unexpected hardware failure preventing suction air flow).

\mypara{Leveraging dense affordances for speed picking.} Our FCNs densely infer affordances for all visible surfaces in the scene, which enables the robot to attempt multiple different suction or grasping proposals (at least 3cm apart from each other) in quick succession until at least one of them is successful (given by immediate feedback from flow sensors or gripper finger width). This improves picking efficiency.

\section{Challenge II: Recognizing Novel Objects with Cross-Domain Image Matching}
\label{sec:recognition}

After successfully grasping an object and isolating it from clutter, the goal of the second step in our system is to recognize the identity of the grasped object.

Since we encounter both known and novel objects, and we have only product images for the novel objects, we address this recognition problem by retrieving the best match among a set of product images. 
Of course, observed images and product images can be captured in significantly different environments in terms of lighting, object pose, background color, post-process editing, etc.
Therefore, we require an algorithm that is able to find the semantic correspondences between images from these two different domains. While this is a task that appears repeatedly in a variety of research topics (\eg domain adaptation, one-shot learning, meta-learning, visual search, etc.), in this paper we refer to it as a \textit{cross-domain image matching} problem (\cite{saenko2010adapting,shrivastava2011data,bell2015learning}).

\subsection{Metric Learning for Cross-Domain Image Matching} 
To perform the cross-domain image matching between observed images and product images, we learn a metric function that takes in an observed image and a candidate product image and outputs a distance value that models how likely the images are of the same object. The goal of the metric function is to map both the observed image and product image onto a meaningful feature embedding space so that smaller $\ell_2$ feature distances indicate higher similarities.  The product image with the smallest metric distance to the observed image is the final matching result.

We model this metric function with a two-stream convolutional neural network (ConvNet) architecture where one stream computes features for the observed images, and a different stream computes features for the product images. We train the network by feeding it a balanced 1:1 ratio of matching and non-matching image pairs (one observed image and one product image) from the set of known objects, and backpropagate gradients from the distance ratio loss (Triplet loss \cite{hoffer2016deep}). This effectively optimizes the network in a way that minimizes the $\ell_2$ distances between features of matching pairs while pulling apart the $\ell_2$ distances between features of non-matching pairs. By training over enough examples of these image pairs across known objects, the network learns a feature embedding that encapsulates object shape, color, and other visual discriminative properties, which can generalize and be used to match observed images of novel objects to their respective product images (Fig. \ref{fig:active-vision}). 

\mypara{Avoiding  metric collapse by guided feature embeddings.}
One issue commonly encountered in metric learning occurs when the number of training object categories is small -- the network can easily overfit its feature space to capture only the small set of training categories, making generalization to novel object categories difficult.  We refer to this problem as metric collapse. To avoid this issue, we use a model pre-trained on ImageNet (\cite{deng2009imagenet}) for the product image stream and train only the stream that computes features for observed images. ImageNet contains a large collection of images from many categories, and models pre-trained on it have been shown to produce relatively comprehensive and homogenous feature embeddings for transfer tasks (\cite{huh2016makes}) -- i.e. providing discriminating features for images of a wide range of objects. Our training procedure trains the observed image stream to produce features similar to the ImageNet features of product images -- i.e., it learns a mapping from observed images to ImageNet features. Those features are then suitable for direct comparison to features of product images, even for novel objects not encountered during training.


\mypara{Using multiple product images.} 
For many applications, there can be multiple product images per object. 
However, with multiple product images, supervision of the two-stream network can become confusing - on which pair of matching observed and product images should the backpropagated gradients be based? \myedit{For example, matching an observed image of the front face of the object against a product image of the back face of the object can easily confuse network gradients.
To solve this problem during training, we add a module called ``multi-anchor switch'' in the network. Given an observed image, this module automatically chooses which ``anchor'' product image to compare against (\ie to compute loss and gradients for) based on $\ell_2$ distance between deep features. We find that allowing the network to select nearest neighbor ``anchor'' product images during training provides a significant boost in performance in comparison to alternative methods like random sampling.}

\subsection{Two Stage Framework for a Mixture of Known and Novel Objects} 

In settings where both types of objects are present, we find that training two different network models to handle known and novel objects separately can yield higher overall matching accuracies. One is trained to be good at ``over-fitting" to the known objects (K-net) and the other is trained to be better at ``generalizing" to novel objects (N-net).

Yet, how do we know which network to use for a given image?  To address this issue, we execute our recognition pipeline in two stages: a ``recollection'' stage that determines whether the observed object is known or novel, and a ``hypothesis'' stage that uses the appropriate network model based on the first stage's output to perform image matching. 


First, the recollection stage infers whether the input observed image from test time is that of a known object that has appeared during training. Intuitively, an observed image is of a novel object if and only if its deep features cannot match to that of any images of known objects. We explicitly model this conditional by thresholding on the nearest neighbor distance to product image features of known objects. In other words, if the $\ell_2$ distance between the K-net features of an observed image and the nearest neighbor product image of a known object is greater than some threshold k, then the observed images is a novel object. Note that the novel object network can also identify known objects, but with lower performance.

In the hypothesis stage, we perform object recognition based on one of two network models: K-net for known objects and N-net for novel objects.  The K-net and N-net share the same network architecture. However, during training the K-net has an ``auxiliary classification" loss for the known objects. \myedit{This loss is implemented by feeding in the K-net features into 3 fully connected layers, followed by an $n$-way softmax loss where $n$ is the number of known object classes. These layers are present in K-net during training then removed during testing.} Training with this classification loss increases the accuracy of known objects at test time to near perfect performance, and also boosts up the accuracy of the recollection stage, but fails to maintain the accuracy of novel objects. On the other hand, without the restriction of the classification loss, N-net has a lower accuracy for known objects, but maintains a better accuracy for novel objects. 

By adding the recollection stage, we can exploit both the high accuracy of known objects with K-net and good accuracy of novel objects with N-net, though incurring a cost in accuracy from erroneous known vs novel classification. We find that this two stage system overall provides higher total matching accuracy for recognizing both known and novel objects (mixed) than all other baselines (Table \ref{table:recognition}).





\section{Experiments}

In this section, we evaluate our affordance-based grasping framework, our recognition algorithm over both known and novel objects, as well as our full system in the context of the Amazon Robotics Challenge 2017. 

\subsection{Evaluating Multi-affordance Grasping}



\begin{figure}[t]
\centering
  \includegraphics[width=\linewidth]{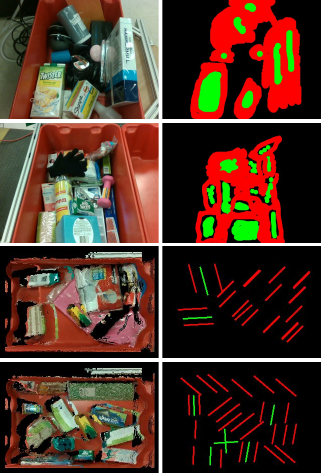}
  \caption{{\bf Images and annotations from the grasping dataset} with labels for suction (top two rows) and parallel-jaw grasping (bottom two rows). Positive labels appear in green while negative labels appear in red.}
  \label{fig:dataset}
  \vspace{-3mm}
\end{figure}

\mypara{Datasets.} To generate datasets for learning affordance-based grasping, we designed a simple labeling interface that prompts users to manually annotate good and bad suction and grasp proposals over RGB-D images collected from the real system. For suction, users who have had experience working with our suction gripper are asked to annotate pixels of suctionable and non-suctionable areas on raw RGB-D images overlooking cluttered bins full of various objects. 
Similarly, users with experience using our parallel-jaw gripper are asked to sparsely annotate positive and negative grasps over re-projected heightmaps of cluttered bins, where each grasp is represented by a pixel on the heightmap and an angle corresponding to the orientation (parallel-jaw motion) of the gripper. On the interface, users directly paint labels on the images with wide-area circular (suction) or rectangular (grasping) brushstrokes. The diameter and angle of the strokes can be adjusted with hotkeys. The color of the strokes are green for positive labels and red for negative labels. Examples of images and labels from this dataset can be found in Fig.~\ref{fig:dataset}. During training, we further augment each grasp label by adding additional labels via small jittering (less than 1.6cm). 
In total, the grasping dataset contains 1837 RGB-D images with pixel-wise suction and grasp labels. 
We use a 4:1 training/testing split of these images to train and evaluate different grasping models.

\myedit{Although this grasping dataset is small for training a deep network from scratch, we find that it is sufficient for fine-tuning our architecture with ResNets pre-trained on ImageNet. An alternative method would be to generate a large dataset of annotations using synthetic data and simulation, as in \cite{mahler2017}.  However, then we would have to bridge the domain gap between synthetic and real 3D data, which is difficult for arbitrary real-world objects (see further discussion on this point in the comparison to DexNet in Table \ref{table:affordance-prediction}). Manual annotations make it easier to embed in the dataset information about material properties which are difficult to capture in simulation (\eg porous objects are non-suctionable, heavy objects are easier to grasp than to suction).}


\mypara{Evaluation.} In the context of our grasping framework, a method is robust if it is able to consistently find at least one suction or grasp proposal that works. To reflect this, our evaluation metric is the precision of inferred proposals versus manual annotations. For suction, a proposal is considered a true positive if its pixel center is manually labeled as a suctionable area (false positive if manually labeled as an non-suctionable area). For grasping, a proposal is considered a true positive if its pixel center is nearby within 4 pixels and 11.25 degrees from a positive grasp label (false positive if nearby a negative grasp label). 

\myedit{We report the precision of our inferred proposals for different confidence percentiles across the testing split of our grasping dataset in Table \ref{table:affordance-prediction}. We compare our method to a heuristic baseline algorithm as well as to a state-of-the-art grasping algorithm Dex-Net \cite{mahler2017,mahler2018} versions 2.0 (parallel-jaw grasping) and 3.0 (suction) for which code is available.} 
\mysecondedit{We use Dex-Net weights pre-trained on their original simulation-based dataset. As reported in \cite{mahler2017,mahler2018,mahler2019learning}, fine-tuning Dex-Net on real data does not lead to substantial increases in performance.}

\begin{table}[h]
  \centering
  \footnotesize
  \setlength{\tabcolsep}{4 pt}
  \caption{Multi-affordance Grasping Performance}
  \begin{tabular}{c|c|c|c|c|c}
    \bf Primitive & \bf Method &\bf Top-1 & \bf Top 1\% &\bf Top 5\% &\bf Top 10\%  \\\hline
    \multirow{2}{*}{Suction} & Baseline & 35.2 & 55.4 & 46.7 & 38.5 \\
    & \myedit{Dex-Net} & \myedit{69.3} & \myedit{71.8} & \myedit{62.5} & \myedit{53.4} \\
    & ConvNet & \bf 92.4 & 83.4 & 66.0 & 52.0 \\\hline 
    \multirow{3}{*}{Grasping} & Baseline & 92.5 & 90.7 & 87.2 & 73.8 \\
     & \myedit{Dex-Net} & \myedit{80.4} & \myedit{87.5} & \myedit{79.7} & \myedit{76.9} \\
    & ConvNet & \bf 96.7 & 91.9 & 87.6 & 84.1 \\\hline 
  \end{tabular}
  \label{table:affordance-prediction}
  \vspace{1mm}
  \\\% precision of grasp proposals across different confidence percentiles.
  \vspace{-3mm}
\end{table}

\myedit{The heuristic baseline algorithm computes suction affordances by estimating surface normal variance over the observed 3D point cloud (lower variance = higher affordance), and computes anti-podal grasps by detecting hill-like geometric structures in the 3D point cloud with shape analysis. Baselines details and code are available on our project webpage \cite{web}. The heuristic algorithm for parallel-jaw grasping was highly fine-tuned to the competition scenario, making it quite competitive with our trained grasping ResNets. We did not compare to the other network architectures for parallel-jaw grasping described in Section \ref{sec:manipulation} since those models could not completely converge during training.}

\myedit{The top-1 proposal from the baseline algorithm performs quite well for parallel jaw grasping, but performs poorly for suction. This suggests that relying on simple geometric cues from the 3D surfaces of objects can be quite effective for grasping, but less so for suction. This is likely because successful suction picking not only depends on finding smooth surfaces, but also highly depends on the mass distribution and porousness of objects -- both attributes of which are less apparent from local geometry alone. Suctioning close to the edge of a large and heavy object may cause the object to twist off due to external wrench from gravity, while suctioning a porous object may prevent a strong suction contact seal. }

\begin{figure}[t]
\centering
  \includegraphics[width=\linewidth]{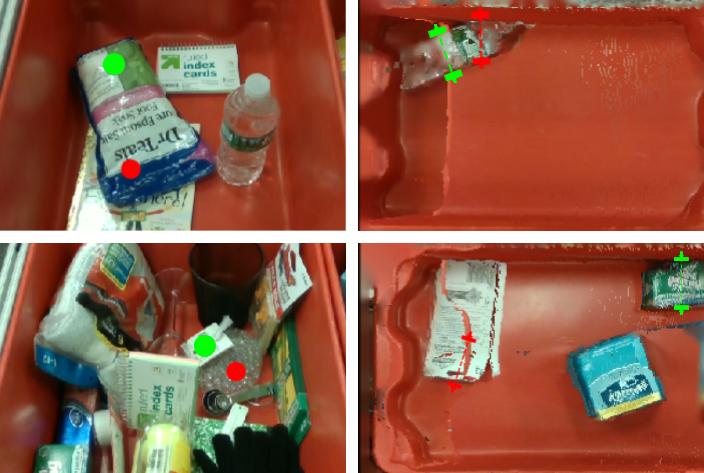}
  \caption{{\bf Common Dex-Net failure modes} for suction (left column) and parallel-jaw grasping (right column). Dex-Net's top-1 predictions are labeled in red, while our method's top-1 predictions are labeled in green. Our method is more likely to predict grasps near objects' center of mass (\eg bag of salts (top left) and water bottle (top right)), more likely to avoid unsuctionable areas such as porous surfaces (\eg mesh bag of marbles(bottom left)), and less susceptible to noisy depth data (bottom right).}
  \label{fig:dex-net-failures}
  \vspace{-3mm}
\end{figure}

\myedit{Dex-Net also performs competitively on our benchmark with strong suction and grasp proposals across top 1\% confidence thresholds, but with more false positives across top-1 proposals. By visualizing Dex-Net top-1 failure cases in Figure \ref{fig:dex-net-failures}, we can observe several interesting failure modes that do not occur as frequently with our method. For suction, there are two common types of failures. The first involves false positive suction predictions on heavy objects. For example, shown in the top left image of Figure \ref{fig:dex-net-failures}, the heavy (~$2kg$) bag of Epsom salt can only be successfully suctioned near its center of mass (\ie near the green circle), which is located towards the bottom of the bag. Dex-Net is expectedly unaware of this, and often makes predictions on the bag but farther from the center of mass (\eg the red circle shows Dex-Net's top-1 prediction). The second type of failure mode involves false positive predictions on unsuctionable objects with mesh-like porous containers. For example, in the bottom left image of Figure \ref{fig:dex-net-failures}, Dex-Net makes suction predictions (\eg red circle) on a mesh bag of marbles -- however the only region of the object that is suctionable is its product tag (\eg green circle).} 

\myedit{For parallel-jaw grasping, Dex-Net most commonly experiences two other types of failure modes. The first is that it frequently predicts false positive grasps on the edges of long heavy objects -- regions where the object would slip due to external wrench from gravity. This is because Dex-Net assumes objects to be lightweight to conform to the payload ($<0.25kg$) of the ABB YuMi robot where it is usually tested. The second failure mode is that Dex-Net often predicts false positive grasps on areas with very noisy depth data. This is likely because Dex-Net is trained in simulation with rendered depth data, so Dex-Net's performance is less optimal without higher quality 3D cameras (\eg industrial Photoneo cameras).} 

\myedit{Overall, these observations show that Dex-Net is a competitive grasping algorithm trained from simulation, but falls short in our application setup due to the domain gap between synthetic and real data.} \mysecondedit{Specifically, the discrepancy between the 90\%+ grasping success achieved by Dex-Net in their reported experiments \cite{mahler2017,mahler2018,mahler2019learning} versus the 80\% on our dataset is likely due to two reasons: our dataset consists of 1) a larger spectrum of objects, e.g., heavier than $0.25kg$; and 2) noisier RGB-D data, i.e., less similar to simulated data, from substantially more cost-effective commodity 3D sensors.}

\mypara{Speed.} Our suction and grasp affordance algorithms were designed to achieve fast run-time speeds during test time by densely inferring affordances over images of the entire scene. Table \ref{table:speed} compares our run-time speeds to several state-of-the-art alternatives for grasp planning. Our numbers measure the time of each FCN forward pass, reported with an NVIDIA Titan X on an Intel Core i7-3770K clocked at 3.5 GHz, excluding time for image capture and other system-related overhead. Our FCNs run at a fraction of the time required by most other methods, while also being significantly deeper (with 101 layers) than all other deep learning methods.

\subsection{Evaluating Novel Object Recognition}

We evaluate our recognition algorithms using a 1 vs 20 classification benchmark. Each test sample in the benchmark contains 20 possible object classes, where 10 are known and 10 are novel, chosen at random. During each test sample, we feed to the recognition algorithm the product images for all 20 objects as well as an observed image of a grasped object. In Table \ref{table:recognition}, we measure performance in terms of average \% accuracy of the top-1 nearest neighbor product image match of the grasped object.
We evaluate our method against a baseline algorithm, a state-of-the-art network architecture for both visual search \cite{bell2015learning} and one-shot learning without retraining \cite{koch2015siamese}, and several variations of our method.  The latter provides an ablation study to show the improvements in performance with every added component:

\begin{table}[t]
  \centering
  \footnotesize
  \setlength{\tabcolsep}{4 pt}
  \caption{Grasp Planning Run-Times (sec.)}
  \begin{tabular}{r|c}
    \bf Method &\bf Time \\\hline
    Lenz et al. \cite{lenz2015rss} & 13.5 \\
    Zeng et al. \cite{zeng2016multi} & 10 - 15 \\
    Hernandez et al. \cite{hernandez2016team} & 5 - 40 \textsuperscript{a} \\
    Schwarz et al. \cite{schwarz2017nimbro} & 0.9 - 3.3 \\
    Dex-Net 2.0 \cite{mahler2017} & 0.8 \\
    Matsumoto et al. \cite{matsumotoend} & 0.2 \\
    Redmon et al. \cite{redmon2015} & 0.07 \\\hline
    Ours (suction) & 0.06 \\
    Ours (grasping) & 0.05$\times{n}$ \textsuperscript{b} \\\hline
  \end{tabular}
  \vspace{1mm}
  \\\textsuperscript{a} times reported from \cite{matsumotoend} derived from \cite{hernandez2016team}.
  \\\textsuperscript{b} $n$ = number of possible grasp angles (in our case $n=$16).
  \label{table:speed}
  \vspace{-3mm}
\end{table}

\mypara{Nearest neighbor}
is a baseline algorithm where we compute features of product images and observed images using a ResNet-50 pre-trained on ImageNet, and use nearest neighbor matching with $\ell_2$ distance. \myedit {For nearest neighbor evaluation, the difference between the matching accuracy for known objects and novel objects reflects the natural difference in distribution of objects in the testing set -- the novel objects are more distinguishable from each other using ImageNet features alone than known objects.} 

\mypara{Siamese network with weight sharing}
is a re-implementation of Bell et al. \cite{bell2015learning} for visual search and Koch et al. \cite{koch2015siamese} for one shot recognition without retraining. We use a Siamese ResNet-50 pre-trained on ImageNet and optimized over training pairs in a Siamese fashion. The main difference between this method and ours is that the weights between the networks computing deep features for product images and observed images are shared.
	
\mypara{Two-stream network without weight sharing}
is a two-stream network, where the networks' weights for product images and observed images are not shared. Without weight sharing the network has more flexibility to learn the mapping function and thus achieves higher matching accuracy. All the later models describe later in this section use this two stream network without weight sharing. 
	
\mypara{Two-stream + guided-embedding (GE)}
includes a guided feature embedding with ImageNet features for the product image stream.  We find this model has better performance for novel objects than for known objects. 

\mypara{Two-stream + guided-embedding (GE) + multi-product-images (MP)}
By adding a multi-anchor switch, we see more improvements to accuracy for novel objects. This is the final network architecture for N-net.

\mypara{Two-stream + guided-embedding (GE) +  multi-product-images (MP) + auxiliary classification (AC)}
By adding an auxiliary classification, we achieve near perfect accuracy of known objects for later models, however, at the cost of lower accuracy for novel objects. This also improves known vs novel (K vs N) classification accuracy for the recollection stage. This is the final network architecture for K-net.

\mypara{Two-stage system}
As described in Section \ref{sec:recognition}, we combine the two different models - one that is good at known objects (K-net) and the other that is good at novel objects (N-net) - in the two stage system. This is our final recognition algorithm, and it achieves better performance than any single model for test cases with a mixture of known and novel objects.

\subsection{Full System Evaluation in Amazon Robotics Challenge}


To evaluate the performance of our system as a whole, we used it as part of our MIT-Princeton entry for the 2017 Amazon Robotics Challenge (ARC), where state-of-the-art pick-and-place solutions competed in the context of a warehouse automation task. Participants were tasked with designing a fully autonomous robot system to grasp and recognize a large variety of different objects from unstructured bins. The objects were characterized by a number of difficult-to-handle properties.  Unlike earlier versions of the competition (\cite{Correll2016}), half of the objects were novel to the robot in the 2017 edition by the time of the competition. The physical objects as well as related item data (i.e. product images, weight, 3D scans), were given to teams just 30 minutes before the competition. While other teams used the 30 minutes to collect training data for the new objects and re-train models, our unique system did not require any of that during those 30 minutes.

\mypara{Setup.}
Our system setup for the competition features several differences. We incorporated weight sensors to our system, using them as a guard to signal stop for grasping primitive behaviors during execution. We also used the measured weights of objects provided by Amazon to boost recognition accuracy to near perfect performance as well as to prevent double-picking. Green screens made the background more uniform to further boost accuracy of the system in the recognition phase. For inferring affordances, Table \ref{table:affordance-prediction} shows that our data-driven methods with ConvNets provide more precise affordances for both suction and grasping than the baseline algorithms. For the case of parallel-jaw grasping, however, we did not have time to develop a fully stable network architecture before the day of the competition, so we decided to avoid risks and use the baseline grasping algorithm. The ConvNet-based approach became stable with the reduction to inferring only horizontal grasps and rotating the input heightmaps. 


\begin{table}[t]
  \centering
  \footnotesize
  \setlength{\tabcolsep}{4.0 pt}
  \caption{Recognition Evaluation (\% Accuracy of Top-1 Match)}
  \begin{tabular}{r|c|c|c|c}
    \bf Method &\bf K vs N &\bf Known &\bf Novel &\bf Mixed \\\hline
    Nearest Neighbor & 69.2 & 27.2 & 52.6 & 35.0 \\
    Siamese (\cite{koch2015siamese}) & 70.3 & 76.9 & 68.2 & 74.2\\\hline  
    Two-stream & 70.8 & 85.3 & 75.1 & 82.2 \\
    Two-stream + GE & 69.2 & 64.3 & 79.8 & 69.0  \\
    Two-stream + GE + MP (N-net) & 69.2 & 56.8 & \bf 82.1 & 64.6 \\
    N-net + AC (K-net) & \bf 93.2 & \bf 99.7 & 29.5 & 78.1  \\\hline
    Two-stage K-net + N-net & 93.2 & 93.6 & 77.5 & \bf 88.6 \\\hline 
  \end{tabular}
  \label{table:recognition}
  \vspace{-5mm}
\end{table}

\mypara{State tracking and estimation.}
We also designed a state tracking and estimation algorithm for the full system in order to perform competitively in the picking task of the ARC, where the goal is to pick \textit{target} objects out of a storage system (\eg shelves, separate work-cells) and place them into specific boxes for order fulfillment. 

\myedit{The goal of our state tracking algorithm is to track all the objects’ identities, 6D poses, amodal bounding boxes, and support relationships in each bin ($bin_{i}$) of the storage system. This information is then used by the task planner during the picking task to prioritize certain pick proposals (close to, or above target objects) over others.
Our state tracking algorithm is built around the assumption that:
1)The state of the objects in the storage system only changes when there is an external force (robot or human) that interacts with the storage system. 
2) We have knowledge of all external interactions in terms of their action type, object category, and specific storage bin.
The action types include: 
\begin{itemize}
\item add ($object_i$, $bin_i$): add $object_i$ to $bin_i$.
\item remove ($object_i$, $bin_i$): remove $object_i$ from $bin_i$.
\item move ($object_i$, $bin_i$): update $object_i$’s location in $bin_i$ (assumes $object_i$  is already in $bin_i$) .
\item touch ($bin_i$): update all object poses in $bin_i$.
\end{itemize}
}

\myedit{
When \textbf{adding} an object into the storage system (\eg during the stowing task), we first use the recognition algorithm described in section \ref{sec:recognition} to identify the object’s class category before placing it into a bin. Then our state tracking algorithm captures RGB-D images of the storage system at time $t$ (before the object is placed) and at time $t+1$ (after the object is placed). The difference between the RGB-D images captured at $t+1$ and $t$ provides an estimate for the visible surfaces of the newly placed object (\ie near the pixel regions with the largest change). 3D models of the objects (either constructed from the same RGB-D data captured during recognition for novel objects or given by another system for known objects) are aligned to these visible surfaces via ICP-based pose estimation (\cite{zeng2016multi}). To reduce the uncertainty and noise of these pose estimates, the placing primitive actions are gently executed -- \ie the robot arm holding the object moves down slowly until contact between the object and storage system is detected with weight sensors, upon which then the gripper releases the object.
}

\myedit{
For the \textbf{remove} operation,  we first verify the object’s identity using the recognition algorithm described in section \ref{sec:recognition}. We then remove the object ID ($object_i$) from the list of tracked objects in $bin_i$.}

\myedit{
The \textbf{move} operation is called whenever the robot attempts to remove an object from a storage bin but fails due to grasping failure. When this operation is called the system will compare the depth images captured before and after the robot’s interaction to identify the moved object’s new point cloud. The system will then re-estimate the object’s pose using the ICP-based method used during the add operation. 
}

\myedit{
The \textbf{touch} operation is used to detect and compensate for unintentional state changes during robot interactions.  This operation is called whenever the robot attempts to add, remove, or move an object in a storage bin.  When this operation is called, the system will compare and compute the correspondence of the color image before and after the interaction using SIFT-flow (\cite{liu2011sift}), ignoring the region of newly added or removed objects. If the difference between the two images is larger than a threshold, we will update each object’s 6D pose by aligning its 3D model to its new corresponding point cloud (obtained from the SIFT-flow) using ICP. 
}


\myedit{
Combined with our affordance prediction algorithm described in section \ref{sec:manipulation}, we are able to label each grasping or suction proposal with corresponding object identities using their tracked 6D poses from the state tracker. The task planner can then prioritize certain grasp proposals (close to, or above target objects) with heuristics based on this information. 
}

\mypara{Results.}
During the ARC 2017 final stowing task, 
we had a 58.3\% pick success with suction, 75\% pick success with grasping, 
and 100\% recognition accuracy during the stow task of the ARC, 
stowing all 20 objects within 24 suction attempts and 8 grasp attempts.
Our system took 1st place in the stowing task, being the only system to have successfully stowed all known and novel objects and to have finished the task well within the allotted time frame.


\myedit{Overall, the pick success rates of all teams in the ARC (62\% on average reported by \cite{morrison2017cartman}) are generally lower than those reported in related work for grasping. We attribute this mostly to the fact that the competition uses bins full of objects that contain significantly more clutter and variety than the scenarios presented in more controlled experiments in prior work. Among the competing teams, we successfully picked the most objects in the Stow and Final Tasks, and our average picking speed was the highest \cite{morrison2017cartman}.}

\mypara{Postmortem.}
\myedit{Our system did not perform as well during the finals task of the ARC due to lack of sufficient failure recovery. On the systems side, the perception node that fetches data from all RGB-D cameras lost connection to one of our RGB-D cameras for recognition and stalled during the middle of our stowing run for the ARC finals, which forced us to call for a hard reset during the competition. The perception node would have benefited from being able to restart and recover from disconnections. On the algorithms side, our state tracking system is particularly sensitive to drastic changes in the state (\ie when multiple objects switch locations), which causes it to lose track without recovery. In hindsight, the tracking would have benefited from some form of simultaneous object segmentation in the bin that works for novel objects and is robust to clutter. Adopting the pixel-wise deep metric learning method of the ACRV team described in \cite{milan2018semantic} would be worth exploring as part of future work.}

\section{Discussion and Future Work}
%

Interest in robust and versatile robotic pick-and-place is almost as old as robotics. Robot grasping and object recognition have been two of the main drivers of robotic research. Yet, the reality in industry is that most automated picking systems are restricted to known objects, in controlled configurations, with specialized hardware.

We present a system to pick and recognize novel objects with very limited prior information about them (a handful of product images). The system first uses an object-agnostic affordance-based algorithm to plan grasps out of four different grasping primitive actions, and then recognizes grasped objects by matching them to their product images.
We evaluate both components and demonstrate their combination in a robot system that picks and recognizes novel objects in heavy clutter, and that took 1st place in the stowing task of the Amazon Robotics Challenge 2017. Here are some of the most salient features/limitations of the system:

\mypara{Object-Agnostic Manipulation.}
The system finds grasp affordances directly in the RGB-D image. This proved faster and more reliable than doing object segmentation and state estimation prior to grasp planning~\cite{zeng2016multi}.
The ConvNet learns the visual features that make a region of an image graspable or suctionable. 
It also seems to learn more complex rules, e.g., that tags are often easier to suction that the object itself, or that the center of a long object is preferable than its ends. 
It would be interesting to explore the limits of the approach. For example learning affordances for more complex behaviors, e.g., scooping an object against a wall, which require a more global understanding of the geometry of the environment.

\mypara{Pick First, Ask Questions Later.}
The standard grasping pipeline is to first recognize and then plan a grasp.
In this paper we demonstrate that it is possible and sometimes beneficial to reverse the order.
Our system leverages object-agnostic picking to remove the need for state estimation in clutter. Isolating the picked object drastically increases object recognition reliability, especially for novel objects.  We conjecture that "pick first, ask questions later" is a good approach for applications such as bin-picking, emptying a bag of groceries, or clearing debris.
It is, however, not suited for all applications -- nominally when we need to pick a particular object. In that case, the described system needs to be augmented with state tracking/estimation algorithms that are robust to clutter and can handle novel objects.

\mypara{Towards Scalable Solutions.}
Our system is designed to pick and recognize novel objects without extra data collection or re-training.
This is a step forward towards robotic solutions that scale to the challenges of service robots and warehouse automation, where the daily number of novel objects 
ranges from the tens to the thousands, making data-collection and re-training cumbersome in one case and impossible in the other.   
It is interesting to consider what data, besides product images, is available that could be used for recognition using out-of-the-box algorithms like ours.

\mypara{Limited to Accessible Grasps.}
The system we present in this work is limited to picking objects that can be directly perceived and grasped by one of the primitive picking motions. 
Real scenarios, especially when targeting the grasp of a particular object, often require plans that deliberately sequence different primitive motions. For example, when removing an object to pick the one below, or when separating two objects before grasping one. 
This points to a more complex picking policy with a planning horizon that includes preparatory primitive motions like pushing whose value is difficult to reward/label in a supervised fashion. Reinforcement learning of policies that sequence primitive picking motions is a promising alternative approach that we have started to explore in \cite{zeng2018learning}.

\mypara{Open-loop vs. Closed-loop Grasping}
Most existing grasping approaches, whether model-based or data-driven are for the most part, based on open-loop executions of planned grasps.
Our system is no different. The robot decides what to do and executes it almost blindly, except for simple feedback to enable guarded moves like move until contact.
Indeed, the most common failure modes are when small errors in the estimated affordances lead to fingers landing on top of an object rather than on the sides, or lead to a deficient suction latch, or lead to a grasp that is only marginally stable and likely to fail when the robot lifts the object.
It is unlikely that the picking error rate can be trimmed to industrial grade without the use of explicit feedback for closed-loop grasping during the approach-grasp-retrieve operation.
Understanding how to make an effective use of tactile feedback is a promising direction that we have started to explore (\cite{donlon2018gelslim, hogan2018tactile}).





\begin{acks}
The authors would like to thank the MIT-Princeton ARC team members for their contributions to this project, and ABB Robotics, Mathworks, Intel, Google, NSF (IIS-1251217 and VEC 1539014/1539099), NVIDIA, and Facebook for hardware, technical, and financial support.
\end{acks}

\bibliographystyle{SageH}
\bibliography{main.bib}

\end{document}